\documentclass[authoryear,review,preprint,12pt,pdf]{elsarticle}
\usepackage[utf8]{inputenc}
\usepackage[T1]{fontenc}
\usepackage{tikz}
\usepackage{makecell}
\usepackage{booktabs}
\usetikzlibrary{shapes.geometric, arrows}
    \usepackage{textcomp}
    \usepackage{tgpagella} % alternative to MinionPro
    \usepackage[centertags, intlimits, sumlimits, namelimits]{amsmath} %  Ajout\'{e} par XB
    \usepackage{amssymb}
\usepackage{natbib}
\usepackage{siunitx} % International units
\usepackage{eurosym}
\catcode`_=12 % To enable the underscore in elsarticle class
\begingroup\lccode`~=`_\lowercase{\endgroup\let~\sb}

\usepackage{breakcites} % Ajouté pour les coupures dans les références
\usepackage{xcolor}
\usepackage{microtype} % ensure optimal character spacing
\usepackage{graphicx}
\usepackage{caption,subcaption}

\usepackage{color}
\usepackage{sidecap}
\usepackage{ifthen}
\usepackage{soul}
    \definecolor{DarkGreen}{rgb}{0.2,0.5,0.2} % to color links in references
\newcommand{\withColorMarking}{true} % change to false if color markings
\ifthenelse{\equal{\withColorMarking}{true}}
  {  }
  {  }
\usepackage{array}
\usepackage{multirow}
\definecolor{DarkBlue}{rgb}{0.2,0.2,0.5} % for citation ref color

  \usepackage[%
       pdftex,%
        colorlinks,%
        citecolor=DarkBlue, % color of references
        hyperindex,%
        plainpages=false,%
        bookmarksopen,%
        bookmarksnumbered %
            ]{hyperref}
\usepackage[english,hyperpageref]{backref}
\newcommand{\mapolicebackref}[1]{
    \hspace*{\fill} \mbox{\textit {\small #1}}
}
\renewcommand*{\backref}[1]{}
\renewcommand*{\backrefalt}[4]{%
\ifcase #1 \mapolicebackref{no citations} % French pas de citations
    \or \mapolicebackref{Cited on page #2} % French : Cité page
    \else \mapolicebackref{#1 citations, pages #2}
\fi
}
\newcommand{\doi}[1]{\url{https://doi.org/#1}} % tell LaTeX to treat DOIs as links
\hypersetup{pdftitle={Deep Learning and Multiple Criteria Optimization: An Application to Multiple Datasets},
            pdfauthor={},
            pdfkeywords={},
            pdfdisplaydoctitle,
            pdfstartview=Fit}
\newtheorem{theorem}{Theorem}
\newtheorem{example}{Example}%[section]
\newtheorem{definition}{Definition}%[section]

\newtheorem{proposition}[theorem]{Proposition}

\newenvironment{proof}[1][Proof]{\noindent\textbf{#1.} }{\ \rule{0.5em}{0.5em}}
\newcommand{\R}{\mathbb{R}}
\newcommand{\N}{\mathbb{N}}

\journal{Annals of Operations Research}
\begin{document}
\begin{frontmatter}

%bibliographystyle{agsm}
%
%\title{How epidemics affect production and markets in time: an optimal control model}
%\title[Production in a pandemic]{Production optimisation in the time of pandemic: an SIS-based optimal control  model with protection effort and cost minimisation}
\title{Mixing Deep Learning and Multiple Criteria Optimization: An  Application to Distributed Learning with Multiple Datasets}

\author[SK]{Davide La Torre}
\ead{davide.latorre@skema.edu}
\author[UM]{Danilo Liuzzi}
\ead{danilo.liuzzi@unimi.it}
\author[SK,BC,SM]{Marco Repetto}
\ead{marco.repetto@skema.edu}
\author[UI]{Matteo Rocca}
\ead{matteo.rocca@uninsubria.it}
\address[SK]{SKEMA Business School, Université Côte d'Azur, France}
\address[UM]{University of Milan, Italy}
\address[UI]{Universit\'a degli Studi dell'Insubria, Italy}
\address[BC]{University of Milan-Bicocca, Milan, Italy}
\address[SM]{Siemens Italy, Milan, Italy}

\begin{abstract}
The training phase is the most important stage during the machine learning process. In the case of labeled data and supervised learning, machine training consists in minimizing the loss function subject to different constraints. In an abstract setting, it can be formulated as a multiple criteria optimization model in which each criterion measures the distance between the output  associated with a specific input and its label. 
Therefore, the fitting term is a vector function and its minimization is intended in the Pareto sense. We provide stability results of the efficient solutions with respect to perturbations of input and output data. 
We then extend the same approach to the case of learning with multiple datasets. 
The multiple dataset environment is relevant when reducing the bias due to the choice of a specific training set. We propose a scalarization approach to implement this model and numerical experiments in digit classification using MNIST data.
\end{abstract}

\end{frontmatter}

\section{Introduction}
It is now accepted that with the term Artificial Intelligence (AI) we identify an interdisciplinary area - which includes biology, computer science, philosophy, mathematics, engineering and robotics, and cognitive science -   focusing on the simulation of human intelligence by means of computer-based machines. 
This is done by training machines that are able to perform tasks normally requiring human intelligence, such as visual perception, speech recognition, decision-making, and translation between languages \cite{wang_barabasi_2021,goel_davies_2011,schank_towle_2000,delahiguera2010}.

Machine Learning (ML) is a branch of AI focusing on algorithms used to learn from data and to make future decisions and predictions \cite{ripley_1996}. There are two main families of ML algorithms: the expression ``supervised learning'' refers to the learning process of an unknown function from labeled training data and based on example input-output pairs. ``Unsupervised learning'', instead, refers to the identification of previously undetected patterns and information in a data set with no pre-existing labels.

Deep Learning (DL) is a subfield of AI and type of ML technique aiming at building systems capable of operating in complex environments \citep{goodfellow2016}. DL systems are based on deep architectures \citep{bottou2007}. 

Fostered by the abundance of data, many recent DL applications require a considerable amount of training. At the same time, local regulations posed significant constraints in terms of data transmission in distributed systems \cite{ahmed2021}. Because of this \cite{konecny2017} proposed the concept of Federated Learning (FL).  FL is a distributed learning methodology allowing model training on a large corpus of decentralized data \cite{bonawitz2019}.
Having distributed data across different nodes requires additional concerns by the Decision Maker (DM), which has to cope with the conflicting objectives of each node as well as potential adversarial attacks \cite{bagdasaryan2020}.

Multiple Criteria Optimization (briefly MOP) is a branch of Operations Research and Decision Making which considers optimization models involving multiple and, in general, conflicting criteria. A growing number of authors have provided advances in this field in the past fifty years and a variety of approaches,  methods, and techniques have been developed for their application in an array of disciplines, ranging from economics to engineering, from finance to management, and many others. Decision making problems with multiple criteria are more complex to be analyzed and they are computationally intensive. However, they usually lead to more informed and better decisions.

In this paper we first formulate the machine training problem as an abstract optimization problem involving a vector-valued functional. The notion of minimization is then intended in the Pareto sense. We provide stability and convergence results of the set of efficient solutions. 
We then extend it to the case of machine training with multiple datasets. 
We present numerical experiments based on scalarization techniques and we validate their performance using digit data from the MINST dataset (see, for instance, \cite{deisenroth_faisal_ong_2020,poole_mackworth_2017,shalev-shwartz_ben-david_2014,barber_2012,jiang_2022,moitra_2018,shah_2020}).
Our results show that the use of multiple criteria optimization methods can also provide a better accuracy of the training algorithm.

The paper is organized as follows. Section \ref{sec:2} presents the key concepts, Deep Learning Architectures and   Multiple Criteria Decision Making. Section \ref{sec:3} presents a vector-valued formulation of machine training with labeled data as well as the main stability properties. Section \ref{sec:4} introduces an extended machine training with multiple data sets. Section \ref{sec:5} and Section \ref{sec:6} present some numerical experiments and, then, Section \ref{sec:8} concludes.

\section{Preliminaries}
\label{sec:2}

\subsection{An Introduction to Deep Learning Architectures}
\label{sec:21}

In general a deep architecture may be defined as: 
$$\mathcal{F} = \{f(\cdot, w), w \in \mathcal{W} \}$$
where $f(\cdot, w)$ is a shallow architecture as, for example, the Perceptron proposed by \cite{rosenblatt1958}. 
As the date of the paper of Rosenblatt suggests, DL has its origin in between the 40s and the 60s in what was called Cybernetics.
Before Rosenblatt, \cite{mcculloch1943} inspired by the spiking behavior of neurons, proposed a system in which binary neurons arranged together were able to do simple logic operations.

It is worth noting that nowadays, neither the Perceptron nor the system proposed by \cite{mcculloch1943} is used in current Artificial Neural Networks (ANNs) configurations.

Modern architectures rely on gradient-based optimization techniques and particularly on Stochastic Gradient Descent (SGD) \citep{saad1998} that requires nonbinary functions.

One of these first architectures trained through gradient-based methods was the Multilayer Perceptron (MLP) \citep{rumelhart1986}.

\begin{figure}[h]
	\begin{center}
		\begin{tabular}{ccc}
			
			\includegraphics[scale=0.4]{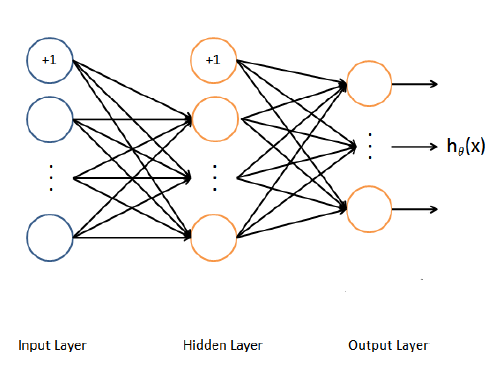}
			
		\end{tabular}
	\end{center}
	\caption{Architecture of the Multilayer Perceptron with a single hidden layer and fully connected nodes}
	\label{DigitsNN}
\end{figure}

The MLP architecture, whose idealized picture is shown in Figure \ref{DigitsNN}, tries to capture the brain's essential functioning by emulating a simple feedforward network of neurons, which are called "Perceptrons" only for historical reason.
In reality, a neuron is an activation function that can either be linear, sigmoidal or piecewise-defined as the rectifier. 

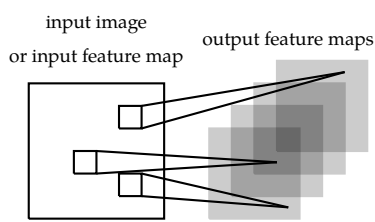
\begin{figure}[h]
	\begin{center}
		\begin{tabular}{ccc}
			\begin{tikzpicture}[thick,scale=0.6, every node/.style={scale=0.6}]
		\node at (1.5,4){\begin{tabular}{c}input image\\or input feature map\end{tabular}};
	
		\draw (0,0) -- (3,0) -- (3,3) -- (0,3) -- (0,0);
		
		\draw (2,2) -- (2.5,2) -- (2.5,2.5) -- (2,2.5) -- (2,2);
		\draw (2,0.5) -- (2.5,0.5) -- (2.5,1) -- (2,1) -- (2,0.5);
		\draw (1,1) -- (1.5,1) -- (1.5,1.5) -- (1,1.5) -- (1,1);
		
		\draw (2.5,2) -- (7,3.25);
		\draw (2.5,2.5) -- (7,3.25);
 
		\draw (2.5,1) -- (5.75,0.25);
		\draw (2.5,0.5) -- (5.75,0.25);
		
		\draw (1.5,1.5) -- (5.5,1.25);
		\draw (1.5,1) -- (5.5,1.25);
		
		\node at (5.75,4){\begin{tabular}{c}output feature maps\end{tabular}};
		
		\draw[fill=black,opacity=0.2,draw=black] (5.5,1.5) -- (7.5,1.5) -- (7.5,3.5) -- (5.5,3.5) -- (5.5,1.5);
		\draw[fill=black,opacity=0.2,draw=black] (5,1) -- (7,1) -- (7,3) -- (5,3) -- (5,1);
		\draw[fill=black,opacity=0.2,draw=black] (4.5,0.5) -- (6.5,0.5) -- (6.5,2.5) -- (4.5,2.5) -- (4.5,0.5);
		\draw[fill=black,opacity=0.2,draw=black] (4,0) -- (6,0) -- (6,2) -- (4,2) -- (4,0);
	\end{tikzpicture}
		\end{tabular}
	\end{center}
	\caption{Interaction between a convolution layer and a pooling layer in a Convolutional Neural Network architecture}
	\label{cnn}
\end{figure}

In image recognition, another well-established architecture is the Convolutional Neural Network (CNN).
The roots of CNN date back to the Neocognitron proposed by \cite{fukushima1982}; however, the first implementation in a supervised learning setting was done by \cite{lecun1990} in digits recognition. 
Instead of the MLP, CNNs do not rely only on fully connected layers. 
Still, they use convolution layers that perform feature extraction through filtering. 
The functioning of a convolution layer is depicted in Figure \ref{cnn}.
After the convolution layer, a second pooling layer is attached that allows for dimensionality reduction.

\begin{figure}[h]
	\begin{center}
		\begin{tabular}{ccc}
			
			\includegraphics{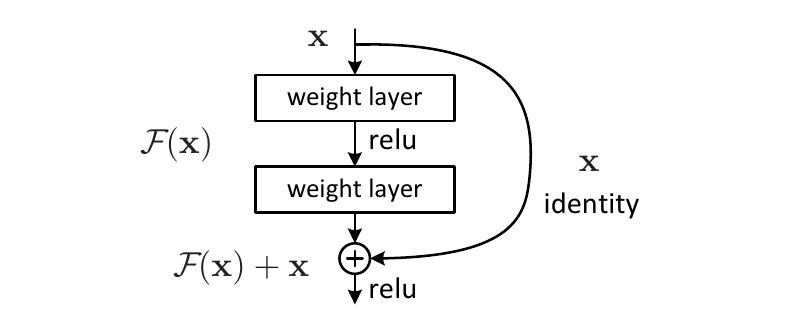}
			
		\end{tabular}
	\end{center}
	\caption{A shortcut connection layer with identity mapping characterizing the Residual Network architecture}
	\label{resnet}
\end{figure}

With the renewed interest in ANNs and DL newer architectures were discovered to counter some of the problems of earlier ANNs, such as the problem of the vanishing gradient.   
Initially proposed by \cite{he2015} ResNet differs from the canonical MLP architecture in that it allows for "shortcut connections" that mitigate the problem of degradation in the case of multiple layers.
Although the usage of shortcut connection is not new in the literature \cite{venables1999}, the key proposal of \cite{he2015} was to use identity mapping instead of any other nonlinear transformation.
Figure \ref{resnet} shows the smallest building block of the ResNet architecture in which both the first and the second layers are shortcutted, and the inputs $x$ are added to the output of the second layer.

The rationale behind ResNet is that by residual learning, the solvers will be able to capture identity mappings that otherwise will be lost in multiple nonlinear layers. 
With shortcuts, identity mapping is achieved by simply annihilating the weights of the input layers that have been shortcutted.
ResNets proved to be a parsimonious yet effective architecture in several image classification tasks \citep{canziani2017a, chen2019, wu2019a}.

\subsection{Basics on Multiple Criteria Optimization}
\label{sec:22}

Multiple Criteria Decision Making (MCDM) involves an optimization model with several and conflicting criteria. Now we briefly recall some basic facts in MCDM that will be useful in the sequel. Given a compact subset $\Omega$ of $\R^n$ and a vector-valued map $J:\Omega\subset\R^n\to\R^p$, $J= (J_1, \ldots , J_p)$ with $J_i:\Omega\subset\R^n\to\R$,  any finite-dimensional MCDM problem can be written:

\begin{eqnarray}
\label{AP}
\min_{x\in \Omega} \ J(x).
\end{eqnarray}

In this paper we suppose that $\R^p$ is ordered by the Pareto cone $\R^p_+$. A point $x\in \Omega$ is said to be Pareto optimal or efficient if $J(x)$ is one of the maximal elements of the set of achievable values in $J(\Omega)$. Thus a point $x$ is Pareto efficient if it is feasible and, for any possible $x'\in X$, $J(x)\le_{\R^p_+} J(x')$ implies $J(x)=J(x')$. In other words, a point $x\in \Omega$ is said to be Pareto efficient if $(J(x) - \R^p_+) \cap J(\Omega) = \{J(x)\}$. We denote by ${\bf Eff}(J)$ the set of efficient points for function $J$. 
We say, instead, that $x\in \Omega$ is weakly Pareto efficient when 
$(J(x) - {\rm int}\, \R^p_+)\cap J(\Omega)= \emptyset$. We denote by ${\rm WEff}(J)$ the set of weakly efficient points for function $J$. Finally, we recall that the point $x$ is properly Pareto  efficient (with respect to $C$) when there exists a cone $C$ with  $\R^p_+ \subseteq {\rm int}\, C$ such that $x$ is Pareto efficient with respect to the cone $C$, i.e. 
\begin{equation}
    (J(x)-C)\cap J(\Omega)= \{J(x)\} 
\end{equation}
We denote by ${\rm PEff}_C(J)$ the set of Pareto properly effient points. Clearly, every properly Pareto efficient point is also Pareto efficient.  For details on the notions of Pareto efficiency one can see e.g. \cite{Sawaragi}. 

One of the most useful techniques  used to reduce a MCDM problem to a single criterion model is the {\it linear scalarization} approach.
In this context a MCDM model can be reduced to a single criterion problem by summing up all criteria with different weights.
The weights express the relative importance of each criterion for the DM. A scalarized version of a MCDM model reads as:

\begin{eqnarray}
\label{SP}
\min_{x\in\Omega} \sum_{i=1}^p \beta_i J_i(x),
\end{eqnarray}
where $\beta= (\beta_1, \ldots , \beta_p)$ is a vector taking values in $\R^p_+$.  Relations between solutions of the scalarized problem (\ref{SP}) and solutions of the vector problem (\ref{AP}) are given in the next proposition. 

\begin{proposition} (see e.g. \cite{Sawaragi})\label{scalarization} The following statements hold true:
\begin{itemize}
    \item [i)] If $\beta \in \R^p_+$, then every solution  of problem (\ref{SP}) is  weakly Pareto efficient for problem (\ref{AP}). If functions $J_i$, $i=1, \ldots , p$ are convex, then the converse holds true, i.e. for every weakly Pareto efficient solution $x$ of  problem (\ref{AP}) there exists a vector $\beta \in \R^p_+$ such that $x$ is a solution of problem  (\ref{SP}). 
    \item [ii)] If $\beta \in {\rm int}\, \R^p_+$ then every solution  of problem (\ref{SP}) is  properly Pareto efficient for problem (\ref{AP}) and hence a Pareto efficient solution. If functions $J_i$, $i=1, \ldots , p$ are convex, then the converse holds true, i.e. for every properly  Pareto efficient solution $x$ of  problem (\ref{AP}) there exists a vector $\beta \in {\rm int}\,\R^p_+$ such that $x$ is a solution of problem  (\ref{AP}).
\end{itemize}

\end{proposition}

Other scalarization methods can be found in the literature that can also be used for non-convex problems. Scalarization can also be applied to problems in which the ordering cone is different than the Pareto one. In this case, one has to rely on the elements of the dual cone to scalarize the problem.

\section{Learning with Labeled Data: A Vector-Valued Formulation} 
\label{sec:3}

Machine training is the essence of supervised machine learning and it is a measure of how well a trained ML model will perform. 
The training phase is crucial for future predictions and in this phase one wants to avoid the problems of {\it overfitting} 
and {\it underfitting}.
We say that a model is {\it well-fitted} when it produces accurate outcomes, something between underfitting and overfitting (see \cite{alpaydin2014,mak_chien_2020,chase_freitag_2019}).

During the training phase, a supervised ML algorithm is run on data for which the target output, known as “labeled” data, is known.
The process involves the minimization of an objective function, called the data-fitting error, over a set of unknown parameters that define the model accuracy.
Over time, as the algorithm learns, the data-fitting error on the training data decreases \citep{blum_hopcroft_kannan_2020,rao_2013,flach_2012}.

Machine training from data consists of finding the optimal model parameters to describe the data. The notion of ``fitting'' provides a measure of how well a model generalizes from given data.

There are different varieties of data-fitting techniques but, in an abstract formulation, most of them can be summarized as follows: 
Given two metric spaces $(X,d^X)$ and $(Y,d^Y)$, a compact set of parameters $\Lambda\subset\R^n$, and a set of input vectors $x_i$ and labels $y_i$, $i=1,\ldots,N$, consider a black box function $f:X\times \Lambda \to Y$ and the following data-fitting/minimization problem:

\begin{equation}
\min_{\lambda\in\Lambda} {\bf DFE}(\lambda) := (d^Y(f(x_1,\lambda),y_1), d^Y(f(x_2,\lambda),y_2), ... , d^Y(f(x_N,\lambda),y_N)) 
\label{DFEProblem}
\end{equation}

The function ${\bf DFE(\lambda)}$ satisfies the following properties:

\begin{itemize}
    \item ${\bf DFE(\lambda)}:\Lambda\to\R^N_+$
    \item if the function $f(x,\cdot)$ is continuous, then ${\bf DFE}$ is continuous over $\Lambda$ and, therefore, ${\bf DFE}$ has at least one global Pareto efficient solution
    \item if there exists $\lambda^*\in\Lambda$ such that ${\bf DFE(\lambda^*)}=0$ then $\lambda^*$ is an ideal - and then efficient - point (In this case $f(x_i,\lambda^*)=y_i$ and this corresponds to the ideal case in which $f(\cdot,\lambda^*)$ maps exactly $x_i$ into $y_i$.)
\end{itemize}

The data-fitting term measures the distance between the empirical values $y_i$ and the theoretical values $f(x_i,\lambda)$ obtained by the black box function if a specific value of $\lambda$ is plugged into it.  

As we can see from this formulation the problem is reduced to the minimization of the vector-valued function ${\bf DFE}(\lambda)$ over the parameters' space $\Lambda$. Depending on the specific function form of $f$ and $d^Y$, the function ${\bf DFE}$ can exhibit different mathematical properties.
The following two examples show how one can obtain classical regression models by specifying the form of $d^Y$ and $f$ and by means of a linear scalarization approach. 

\begin{example}
Let us suppose that $f(x,\lambda)=\lambda\cdot x$,  
$d^Y(f(x_i,\lambda),y_i) = (\lambda\cdot x_i - y_i)^2$,
and scalarization coefficients are $\beta_i={1\over N}$, $i=1...N$. 
Then the scalarization of the above model (\ref{DFEProblem}) takes the form:
\begin{equation}
\min_{\lambda\in\Lambda} {\bf DFE} := {1\over N} \sum_{i=1}^N (\lambda\cdot x_i - y_i)^2
\end{equation}
which coincides with the mean squared error.
\end{example}

\begin{example}
Suppose that $y_i\in\{-1,1\}$ and $d^Y(f(x_i,\lambda),y_i) = \phi(f(x_i,\lambda) y_i)$ where $\phi(u)= \ln(1+e^{-u})$ and $\beta_i={1\over N}$. 
Then the scalarization of the above model (\ref{DFEProblem}) takes the form
\begin{equation}
\min_{\lambda\in\Lambda} {\bf DFE} := {1\over N} \sum_{i=1}^N \ln(1+e^{-f(x_i,\lambda) y_i})
\end{equation}
which coincides with the logistic regression model.
\end{example}

\begin{example}
Suppose that $y_i\in\{0,1\}$ then $$d^Y(f(x_i,\lambda),y_i) = - \sum_{i=1}^{N} \left[ y_i \log(f(x_i,\lambda)) + (1-y_i)\log(1-f(x_i,\lambda)) \right]$$
and scalarization coefficients are $\beta_i={1\over N}$, $i=1...N$. 
Then the above problem (\ref{DFEProblem}) takes the form:
\begin{equation}
\min_{\lambda\in\Lambda} {\bf DFE} := -{1\over N} \sum_{i=1}^N \left[ y_i \log(f(x_i,\lambda)) + (1-y_i)\log(1-f(x_i,\lambda)) \right]  
\end{equation}
which coincides with the Binary Cross Entropy loss with reduction.
\end{example}

The following result states a stability result of the function ${\bf DFE}$ with respect to perturbation of the label set. 

\begin{proposition}
\label{prop1}
Let $\{(x_i,y_i)\}$ and $\{(x_i,\tilde y_i)\}$ be two data sets with the same numerosity $N\in\N$, and let ${\bf DFE(\lambda)}$ and ${\bf \tilde DFE(\lambda)}$ be the two corresponding fitting functions. Then
\begin{equation}
\|{\bf DFE(\lambda)} - {\bf {\tilde DFE}(\lambda)}\|_2\le \sqrt{\sum_{i=1}^N d^Y(y_i,\tilde y_i)^2} 
\end{equation}
\end{proposition}

The following result, instead, provides a condition for the problem stability with respect to perturbation of the input data. 

\begin{proposition}
\label{prop2}
Let $\{(x_i,y_i)\}$ and $\{(\tilde x_i,y_i)\}$ be two data sets with the same numerosity $N\in\N$, and let ${\bf DFE(\lambda)}$ and ${\bf \tilde DFE(\lambda)}$ be the two corresponding fitting functions. 
Let us suppose that $f(x, \lambda)$ is Lipschitz with respect to $x$, that is there exist $K$ such that $d^Y(f(a,\lambda),f(b,\lambda))\le K d^X(a,b)$ for any $a,b\in X$ and $\lambda \in \Lambda$. 
Then
\begin{equation}
\|{\bf DFE(\lambda)} - {\bf {\tilde DFE}(\lambda)}\|_2\le K \sqrt{\sum_{i=1}^N  d^X(x_i,\tilde x_i)^2} 
\end{equation}
\end{proposition}

\begin{proposition}
\label{convergence} 
Let $(x_i^n, y_i^n)$ be  sequences in $X \times Y$  converging to $(x_i,y_i)$  in the $d_{X\times Y} = d_X + d_Y$ metric such that $f(x_i^n, \lambda)$ converges  to $f( x_i, \lambda)$, uniformly with respect to $\lambda \in \Lambda$, $i=1, \ldots , N$. Assume $f(x, \cdot)$ is continuous and let
\begin{equation}
{\bf DFE}_n(\lambda) := (d^Y(f(x_1^n,\lambda),y_1^n), d^Y(f(x_2^n,\lambda),y_2^n), ... , d^Y(f(x_N^n,\lambda),y_N^n)) 
\label{DFEProblem-n}
\end{equation}

\begin{itemize}
    \item [i)] Let $\lambda_n \in {\rm WEff}({\bf DFE}_n)$. Then there exists a subsequence $\lambda_{n_k}$ converging to $\bar \lambda \in \Lambda$ such that $\bar \lambda \in {\rm WEff}({\bf DFE})$.
    \item [ii)] Let $\lambda_n \in {\rm PEff}_C({\bf DFE}_n)$, with $\R^p_+ \subseteq{\rm int}\, C$. Then there exists a subsequence $\lambda_{n_k}$ converging to $\bar \lambda \in \Lambda$ such that $\bar \lambda \in {\rm Eff}({\bf DFE})$.
   
\end{itemize}
\end{proposition}

%For $v=(v_1, \ldots , v_n) \in \R^n$ now we set 
%\begin{equation}
%\|v\|_1= |v_1|+ \cdots |v_n|
%\end{equation}
For two subsets of $\Lambda$,  $A$ and $C$, we set
\begin{equation}
e(A, C)={\rm sup}_{a \in A}d(a, C)
\end{equation}
with $d(a,C)= {\rm inf}_{c \in C}\|a-c\|$.
In the following, for simplicity sake,  let  $z_i = (x_i, y_i)\in Z=X\times Y$,  $z=(z_1, \ldots , z_N)\in Z^N$, $z^0=(z_1^0, \ldots ,z_N^0)$. 
We assume $Z$ is a metric space with distance $d^Z=d^X+d^Y$. \ \\ Let  $g_i(\lambda, z_i)=d_i(f(x_i, \lambda), y_i)$ , $i =1, \ldots , N$ and 
$$
g(\lambda)= (g_1(\lambda, z_1), \ldots , g_N(\lambda), z_N)={\bf DFE(\lambda)}
$$
Minimizing $g$ clearly means minimizing ${\bf DFE}(\lambda)$, with data given by the vector $z$. We denote by ${\rm Eff}_z({\bf DFE})$ the set of efficient solutions with data set given by $z$. 

\begin{definition} 
\label{sgdp} (see e.g. \cite{Li}) 
Let $f: \Lambda \to \R$. We say that $\lambda_0 \in \Lambda$ is an isolated minimizer of  order $\alpha >0$ and constant $h>0$  when  for every $\lambda \in \Lambda$ it holds
\begin{equation}\label{isolated}
f(\lambda)-f(\lambda_0)\ge  h\|\lambda - \lambda_0\|^{\alpha}
\end{equation}
We say that $\lambda_0 \in \Lambda$ is a local isolated minimizer of  order $\alpha >0$ and constant $h>0$ when (\ref{isolated}) holds for $\lambda$ in a neighborhood of $\lambda_0$. 
\end{definition}

Let $\beta_i\in (0, 1]$, $i=1, \ldots , N$  with $\sum_{i=1}^N\beta_i=1$ and consider function $l(\lambda, z)= \sum_{i=1}^N \beta_i g_i(\lambda , z_i)$ . Denote by $S_z^l$ the set minimizers of $l(\cdot , z)$ over $\Lambda$. It is well known that $S_z^l \subseteq {\rm Eff}_z({\bf DFE})$ where ${\rm Eff}_z({\bf DFE})$ denotes the set of efficient points of {\bf DFE} with data set given by $z$ (see Proposition (\ref{scalarization})). 

\begin{proposition}
\label{estimation}
Assume that 
\begin{itemize}
\item [i)] for some choice of scalars $\beta_i\in (0, 1]$, $i=1, \ldots , N$  with $\sum_{i=1}^N\beta_i=1$ there exists a point $\lambda(z^0) \in \Lambda$ that is an  isolated minimizer of  order $\alpha >0$ and constant $h>0$ for  $l(\cdot, z^0)$. 
\item[ii)] for any $\lambda \in \Lambda$, each  $g_i (\lambda, \cdot)$ is H\"older of order $\delta>0$ on $Z$ with constant $m>0$, i.e. for any $z_i^1, \ z_i^2 \in Z$ it holds 
\begin{equation}
|g_i(\lambda, z_i^1)-g_i(\lambda, z_i^2)|\le m d^Z (z_i^1, z_i^2)^{\delta}
\end{equation}
\end{itemize}
Then it holds
\begin{equation}
e(S_z^l, S_{z^0}^l)\le \left( \frac{2m}{h}\right)^{1/\alpha}\left(\sum_{i=1}^N d^Z(z_i, z_i^0)^{\delta}\right)^{1/\alpha}
\end{equation}
Consequently, there exists $\lambda(z) \in {\rm Eff}_z({\bf DFE})$ such that 
\begin{equation}
d(\lambda(z), {\rm Eff}_{z^0}({\bf DFE}))\le \left( \frac{2m}{h}\right)^{1/\alpha}\left(\sum_{i=1}^N d^Z(z_i, z_i^0)^{\delta}\right)^{1/\alpha}.
\end{equation}
\end{proposition}

Proposition (\ref{estimation}) admits a local version presented in the following result.  

\begin{proposition}
\label{sta}
Assume that 
\begin{itemize}
\item [i)] for some choice of scalars $\beta_i\in (0, 1]$, $i=1, \ldots , N$  with $\sum_{i=1}^N\beta_i=1$ there exists a point $\lambda(z^0) \in \Lambda$ that is a local   isolated minimizer of  order $\alpha >0$ and constant $h>0$ for  $l(\cdot, z^0)$. 
\item[ii)] for any $\lambda \in \Lambda$, each  $g_i (\lambda, \cdot)$ is H\"older of order $\delta>0$ on $Z$ with constant $m>0$. 
\item [iii)] for $z \to z^0$, $g_i(\lambda, z)\to g_i( \lambda, z^0)$ uniformly with respect to $\lambda$ in a neighborhood of $\lambda (z^0)$. 
\end{itemize}
Then there exists a neighborhood $U$ of $z^0$ such that for every $z \in U$ one can find $\lambda (z) \in {\rm Eff}_z({\bf DFE})$ such that 
\begin{equation}
d(\lambda(z), {\rm Eff}_{z^0}({\bf DFE}))\le \left( \frac{2m}{h}\right)^{1/\alpha}\left(\sum_{i=1}^N d^Z(z_i, z_i^0)^{\delta}\right)^{1/\alpha}.
\end{equation}
\end{proposition}

\section{Multiple datasets}
\label{sec:4}

Now let us consider the case in which we have different data sets, $\Gamma_1$, $\Gamma_2$, ... $\Gamma_M$, each of them with dataset numerosity $s_i$.
In this context we want to learn from different datasets simultaneously. This approach allows to reduce the bias in the training process due to the choice of a particular dataset. It is worth mentioning that the stability results proved in the previous sections can be easily extended to this context. The training process in this context reads as

\begin{equation}
\min_{\lambda\in\Lambda}  ({\bf DFE}_{s_1}^1(\lambda),...,({\bf DFE}_{s_M}^M(\lambda)) 
\label{model}
\end{equation}
where ${\bf DFE}_{s_1}^1:\Lambda\to\R^{s_1}$, ${\bf DFE}_{s_2}^2(\lambda):\Lambda\to\R^{s_2}$, ... ${\bf DFE}_{s_M}^M(\lambda):\Lambda\to\R^{s_M}$ are defined as:

\begin{equation}
{\bf DFE}_{s_1}^1(\lambda) = (d^Y(f(x_i,\lambda),y_i), (x_i,y_i)\in\Gamma_1))\in\R^{s_1}
\end{equation}
\begin{equation}
{\bf DFE}_{s_2}^2(\lambda) = (d^Y(f(x_i,\lambda),y_i), (x_i,y_i)\in\Gamma_2))\in\R^{s_2}
\end{equation}
and 
\begin{equation}
{\bf DFE}_{s_M}^M(\lambda) = (d^Y(f(x_i,\lambda),y_i), (x_i,y_i)\in\Gamma_M))\in\R^{s_M}
\end{equation}

One possible way to solve the above model is to rely on the linear scalarization approach. If we denote by ${\bf \beta_i}\in \R^{s_i}$, $i=1, \ldots , M$, the weights associated with each criterion, the scalarized model reads as
\begin{equation}
\label{eq:sc}
\min_{\lambda\in\Lambda}  {\bf \beta_1} {\bf DFE}_{s_1}^1(\lambda)  + ... + {\bf \beta_M} {\bf DFE}_{s_M}^M(\lambda) 
\end{equation}

It is clear that the theory proved in the previous section applies to 
Eq. (\ref{eq:sc}) as well and the stability results can be reformulated  in terms of set-to-set distances. 

\section{Numerical experiments}
\label{sec:5}

In this section we describe a computational experiment using digit recognition and a set ANN architectures. 
As it is usually the case for image recognition, the input layer corresponds to a vectorized form of the training image: a $n\times m$ pixels grayscale image can be digitalized as a $n\times m$ matrix and then transformend into a vector of dimension $nm\times 1$. 

\begin{figure}[t]
	\begin{center}
		\begin{tabular}{ccc}
		\includegraphics[scale=0.4]{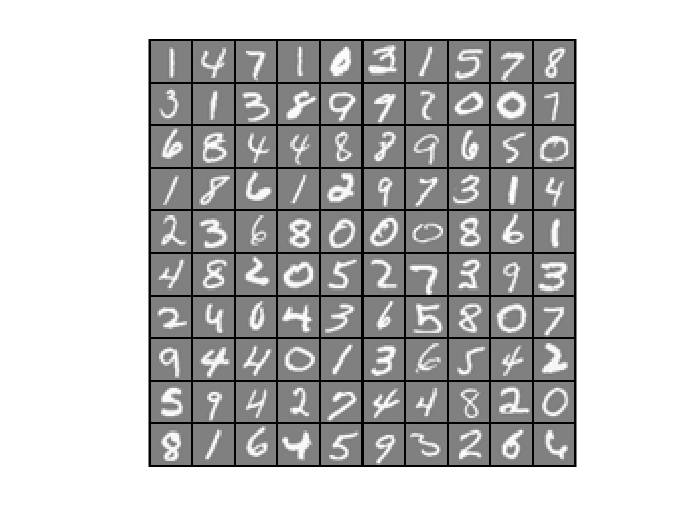}
		\end{tabular}
	\end{center}
	\caption{Handwritten digits from the MNIST database}
	\label{Digits}
\end{figure}

The training dataset is a subset $\Gamma$ of the widely known MNIST database which has been used to test several families of ML classification algorithms \footnote{Available at http://yann.lecun.com/exdb/mnist/}. Some sample data are shown in Figure \ref{Digits}.

A digit is a 20x20 pixels image, whose corresponding matrix contains numbers between $0$ and $255$, proportional to the brightness of the pixel.

For the analysis purpose, the database $\Gamma$ has been divided in three equally sized  subsets, $\Gamma_1$,   $\Gamma_2$ and $\Gamma_3$ ($s_1=s_2=s_3$).
$\Gamma_1$ contains a third of the untouched original data, while $\Gamma_2$ and $\Gamma_3$ contain, respectively, data that have been augmented with a Gaussian noise with zero mean and standard deviations $\sigma_2$ and $\sigma_3$.

A scalarization approach of the above vector problem (\ref{model}) leads to the following:
\begin{equation}
\min_{\lambda\in\Lambda}  \beta_1 {\bf DFE}_{s_1}^1(\lambda) +\beta_2 {\bf DFE}_{s_2}^2(\lambda)+\beta_3 {\bf DFE}_{s_3}^3(\lambda)
%+ \beta_{4} SP(\lambda)
\label{model_example}
\end{equation}

where  $\beta_i$ is the weight associated with the $i$-th term, while ${\bf DFE}_{s_i}^i$ refers to the data fitting function defined using the dataset $\Gamma_i$. Here, with some abuse of notation, we still identify the scalarized  
data fitting terms over each dataset with the expressions ${\bf DFE}_{s_i}^i(\lambda)$.  

To test our approach we choose $\beta={1\over 3}$, which corresponds to case of no database splitting, and we perturb each weight by an $\epsilon$ for each architecture. The resulting scalarized loss function is therefore the following:

\begin{equation}
\min_{\lambda\in\Lambda}  \left(\frac{1}{3} + \epsilon\right) {\bf DFE}_{s_1}^1(\lambda)+ \left(\frac{1}{3} - \frac{\epsilon}{2}\right) {\bf DFE}_{s_2}^2(\lambda)+ \left(\frac{1}{3} - \frac{\epsilon}{2
}\right) {\bf DFE}_{s_3}^3(\lambda)
%+ \beta_{4} SP(\lambda)
\label{loss}
\end{equation}

Of course when $\epsilon=0$ we obtain the basic formulation. For comparable results, all the ANNs architectures have been optimized with SGD with the same learning rate, and weights were initialized pseudo-randomly under the same seed.

\subsection{The case of the Multilayer Perceptron}
\label{sec:7}

This section describes a computational experiment by employing the MLP architecture. 
In this context the fitting function ${\bf DFE}_{s_i}^i$ is given by the following:

\begin{equation}
{\bf DFE}_{s_i}^i(\lambda) = \frac{1}{s_i}\sum_{j=0}^{s_i}\sum_{k=1}^{K}[y_j^{(k)}\,\log((h_{\lambda}(x_j))_k) + (1-y_j^{(k)})\,\log(1-(h_{\lambda}(x_j))_k)]
\label{cost_f_i}
\end{equation}

where $s_i$ is the cardinality of $\Gamma_i$. The hypothesis function $(h_{\lambda}(x_j))_k$ is based on forward propagation: each unit in the second and the third layer processes the linear combination ($\lambda^T x_j$) of its incoming signals via a sigmoid function:
\begin{equation}
h_{\lambda}(x_j)= \frac{1}{1+e^{\lambda^T x_j}}. 
\end{equation}
The index $k=1..K$ represent the $k^{th}$ label. The matrices $\lambda^{(1)}$ and $\lambda^{(2)}$ represent the forward propagation from layer $1$ to layer $2$ and from layer $2$ to layer $3$, respectively.

In  order to explore a neighborhood of this configuration we
perturbed the weight $\beta_i$ with a small amount $\epsilon$. 
In this numerical example we use a MLP with an input layer, a hidden layer and an output layer.  As the dataset contains $20\times 20$ pixels images this implies that we need an input layer with $N=400$ nodes. The output layer has $K=10$ nodes and the hidden layer has been chosen to have $H=25$ nodes. 

%The hypothesis function $(h_{\lambda}(x_j))_k$ is based on forward propagation: each unit in the second and the third layer processes the linear combination ($\lambda^T x_j$) of its incoming signals via a sigmoid function:
%\begin{equation}
%h_{\lambda}(x_j)= \frac{1}{1+e^{\lambda^T x_j}}. \label{sigmoid}  
%\end{equation}
%The index $k=1..K$ represent the $k^{th}$ label. 
 
\begin{figure}[h]
	\begin{center}
		\begin{tabular}{ccc}
			
			\includegraphics[scale=0.4]{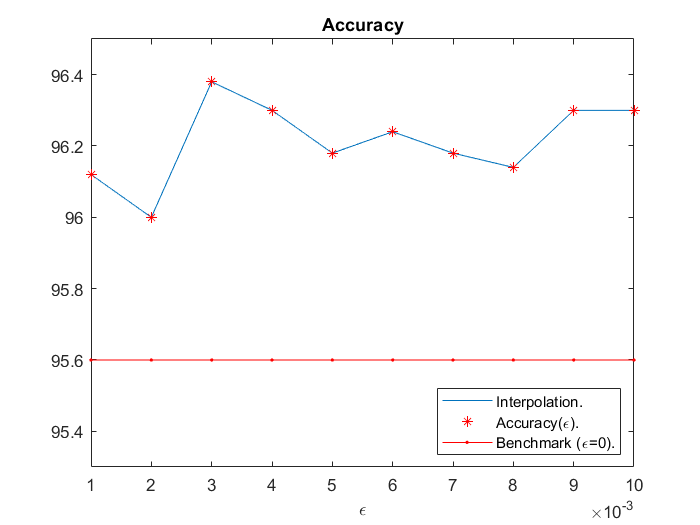}
			
		\end{tabular}
	\end{center}
	\caption{Accuracy on the MNIST dataset of the Multilayer Percepron with different values of $\epsilon$ perturbation (blue line), against the non perturbed benchmark (red line)}
	\label{acc_eps}
\end{figure}

Figure \ref{acc_eps} shows how the accuracy reacts to change in the values of the "perturbation" parameter $\epsilon$. Uniformly varying $\epsilon$ within the range $[0.001,0.01]$ it is possible to corroborate the previous results on the improvement of the accuracy: the different accuracy levels are compared to the benchmark one obtained with $\epsilon=0$.

\subsection{The case of ResNet}
\label{sec:71}
The second numerical experiment employs a more recent ANN architecture, the ResNet.

\begin{figure}[h]
	\begin{center}
		\begin{tabular}{ccc}
			
		\begin{tikzpicture}[node distance=2cm, scale=0.6, every node/.style={scale=0.6}]
		\tikzstyle{empty} = [circle, minimum width=0cm, minimum height=0cm,text centered, draw=white]
        \tikzstyle{layer} = [rectangle, minimum width=3cm, minimum height=1cm,text centered, draw=black]
        \tikzstyle{identity} = [circle, minimum width=0.5cm, minimum height=0.5cm,text centered, draw=black]
        \node (e1) [empty] {};
         \node (l1) [layer, below of=e1] {20x20 linear, 64};
        \node (l2) [layer, below of=l1] {64 linear, 64};
        \node (i1) [identity, below of=l2] {+};
        \node (l3) [layer, below of=i1] {64 linear, 10};
        \node (e2) [empty, below of=l3] {};
        \draw[->] (e1) -> (l1);
        \draw[->] (l1) -> (l2) node[midway,left] {Relu};
        \draw[->] (l2) -> (i1) node[midway,left] {Relu};
        \draw[->] (i1) -> (l3);
        \draw[->] (l2.north) to[out=20, in=-10, looseness=5] (i1.east);
        \draw[->] (l3) -> (e2);
        \end{tikzpicture}
		\end{tabular}
	\end{center}
	\caption{Residual Network architecture employed in the numerical experiment}
	\label{full_resnet}
\end{figure}
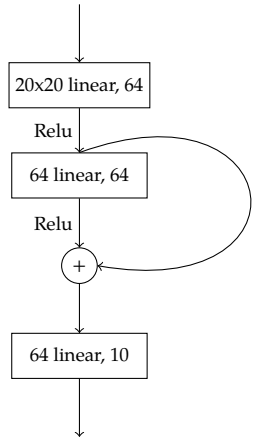

The configuration used in the numerical experiment is reported in figure \ref{full_resnet}. 
The flattened input image passes through the first and second layers with the rectifier activation function, that is:
\begin{equation}
h_{\lambda}(x_j)= \max \{0, \lambda^T x_j\} 
\end{equation}
The shortcut is placed in the output of layer one and added to the output of layer two. 

\begin{figure}[h]
	\begin{center}
		\begin{tabular}{ccc}
			
			\includegraphics[scale=0.4]{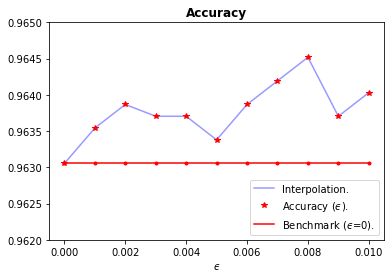}
			
		\end{tabular}
	\end{center}
	\caption{Accuracy on the MNIST dataset of the Residual Network with different values of $\epsilon$ perturbation (blue line), against the non perturbed benchmark (red line)}
	\label{resnet_result}
\end{figure}

To keep the numerical experiment comparable with what was proposed in the previous section, we adopted the same cost function as in \ref{cost_f_i}.
Figure \ref{resnet_result} shows the accuracy as the perturbation parameter $\epsilon$ varies.

The results corroborate the theoretical findings of the previous sections.

\subsection{The case of Convolutional Neural Networks}
\label{sec:72}
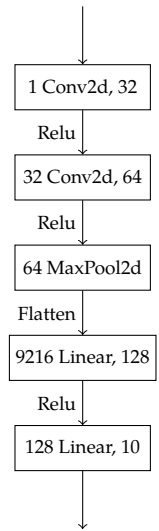
\begin{figure}[h]
	\begin{center}
		\begin{tabular}{ccc}
			
		\begin{tikzpicture}[node distance=2cm, scale=0.6, every node/.style={scale=0.6}]
		\tikzstyle{empty} = [circle, minimum width=0cm, minimum height=0cm,text centered, draw=white]
        \tikzstyle{layer} = [rectangle, minimum width=3cm, minimum height=1cm,text centered, draw=black]
        \tikzstyle{identity} = [circle, minimum width=0.5cm, minimum height=0.5cm,text centered, draw=black]
        \node (e1) [empty] {};
         \node (l1) [layer, below of=e1] {1 Conv2d, 32};
        \node (l2) [layer, below of=l1] {32 Conv2d, 64};
        \node (l3) [layer, below of=l2] {64 MaxPool2d};
        \node (l4) [layer, below of=l3] {9216 Linear, 128};
        \node (l5) [layer, below of=l4] {128 Linear, 10};
        \node (e2) [empty, below of=l5] {};
        \draw[->] (e1) -> (l1);
        \draw[->] (l1) -> (l2) node[midway,left] {Relu};
        \draw[->] (l2) -> (l3) node[midway,left] {Relu};
        \draw[->] (l3) -> (l4) node[midway,left] {Flatten};
        \draw[->] (l4) -> (l5) node[midway,left] {Relu};
        \draw[->] (l5) -> (e2) node[midway,left] {};
       \end{tikzpicture}
		\end{tabular}
	\end{center}
	\caption{Convolutional Neural Network architecture employed in the numerical experiment}
	\label{full_convnet}
\end{figure}

For the last numerical example, the architecture used is the Convolutional Neural Network (CNN). In our experiment, we used the architecture depicted in Figure \ref{full_convnet}.

\begin{figure}[h]
	\begin{center}
		\begin{tabular}{ccc}
			
			\includegraphics[scale=0.4]{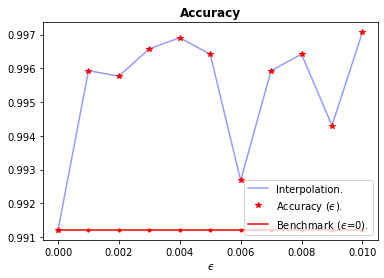}
			
		\end{tabular}
	\end{center}
	\caption{Accuracy on the MNIST dataset of the Convolutional Neural Network with different values of $\epsilon$ perturbation (blue line), against the non perturbed benchmark (red line)}
	\label{convnet_result}
\end{figure}

Figure \ref{convnet_result} shows the change in accuracy with the epsilon perturbation on the betas with and without regularization. 

\section{Testing on the validation set}
\label{sec:6}
The results of the previous numerical experiments have been analyzed only from the perspective of the training process.
However, in the general empirical setting, the performance is taken from a set, or multiple sets, of data that is not used in training.
This set of data is called the validation set and serves a different role from the test set.
The necessity of a validation set is twofold. 
First, the model performance on a validation set is a rough proxy of the model's generalization error \citep{friedman2017}. 
Second, the validation set is also used to check the goodness of possible hyperparameters configurations \citep{goodfellow2016}.
In our case, $\epsilon$ should be treated as a hyperparameter.
Therefore our empirical setting uses the results derived from the best neural network architecture out of the three used in the numerical experiments.
Then we select the model with the value of $\epsilon$ achieving the best performance in the training set and compare it with the model's accuracy in the validation set with respect to the benchmark in which $\epsilon =0$. 

\begin{table}
\centering
\caption{Accuracy on the training and validation sets, with $\epsilon=0.0$ and $\epsilon=0.01$}
\label{table:network}
\resizebox{0.4\columnwidth}{!}{%
\begin{tabular}{@{}lcc@{}}
\toprule
\textbf{$\epsilon$} & \textbf{Training set} & \textbf{Validation set} \\ \midrule
$0.0$ & $0.9912$ & $0.9668$ \\
$0.01$ & $0.9971$ & $0.9673$ \\ \bottomrule
\end{tabular}
}
\end{table}

\section{Conclusion}
\label{sec:8}
We have formulated the machine training model as a vector-valued optimization problem in which each criterion measures the distance between the output value associated with an input value and its label. We have proved stability results for this problem. 
We have then considered the case of multiple datasets. In this case, the training can be split over each dataset simultaneusly and this leads to an extended multicriteria setting. We have applied this model to the case of Multilayer Perceptron, Deep Residual Network and Convolutional Neural Network via scalarization approach. 
Our numerical simulation shows that the adoption of multicriteria techniques can not only provide a general framework to contextualize the machine training with multiple datasets but it can also provide a better accuracy and performance if the right choice of the weights is implemented. 
Future works include the implementation of more advanced multicriteria techniques. 

\bibliographystyle{apalike}
{\footnotesize
\bibliography{paper}}

\section{Technical Appendix}

\begin{proof}
This is the proof of Proposition \ref{prop1}. By computing, we get:
\begin{equation}
\|{\bf DFE(\lambda)} - {\bf \tilde DFE(\lambda)}\|_2^2 = 
\end{equation}
$$
\|(d^Y(f(x_1,\lambda),y_1), ... , d^Y(f(x_N,\lambda),y_N))-(d^Y(f(x_1,\lambda),\tilde y_1), ..., d^Y(f(x_N,\lambda), \tilde y_N)) \|_2^2 =
$$
$$
(d^Y(f(x_1,\lambda),y_1)-d^Y(f(x_1,\lambda),\tilde y_1))^2 + ... + (d^Y(f(x_N,\lambda),y_N)-d^Y(f(x_N,\lambda),\tilde y_N))^2\le 
$$
$$
d^Y(y_1,\tilde y_1)^2 + ... + d^Y(y_N,\lambda y_N)^2
$$
where the last inequality follows from the distance property:
$$
|d^Y(a,b)-d^Y(b,c)|^2 \le d^Y(a,c)^2
$$
\end{proof}

\begin{proof}
This is the proof of Proposition \ref{prop2}. By computing, we get:
\begin{equation}
\|{\bf DFE(\lambda)} - {\bf \tilde DFE(\lambda)}\|_2^2 = 
\end{equation}
$$
\|(d^Y(f(x_1,\lambda),y_1), ... , d^Y(f(x_N,\lambda),y_N))-(d^Y(f(\tilde x_1,\lambda),y_1), ..., d^Y(f(\tilde x_N,\lambda),y_N)) \|_2^2 =
$$
$$
(d^Y(f(x_1,\lambda),y_1)-d^Y(f(\tilde x_1,\lambda), y_1))^2 + ... + (d^Y(f(x_N,\lambda),y_N)-d^Y(f(\tilde x_N,\lambda),y_N))^2\le 
$$
$$
K^2 d^X(x_1,\tilde x_1)^2 + ... + K^2 d^Y(x_N,\tilde x_N)^2
$$
and now the thesis easily follows.
\end{proof}

\begin{proof}
This is the proof of Proposition \ref{convergence}.
\begin{itemize}
    \item [i)] Since $\Lambda$ is compact, $\lambda_n$ admits a subsequence $\lambda_{n_k}$ converging to $\bar \lambda \in \Lambda$. Since $\lambda_{n_k}\in {\rm WEff}({\bf DFE}_{n_k})$, we have
     \begin{equation}
    {\bf DFE}_{n_k}(\lambda)- {\bf DFE}_{n_k}(\lambda_{n_k}) \not \in - {\rm int}\, \R^N, \ \forall \lambda \in \Lambda
    \label{weff}    
    \end{equation}
    Further we have
    \begin{equation}
    \begin{split}
       & \|{\bf DFE}_{n_k}(\lambda_{n_k})- {\bf DFE}(\bar \lambda\|_2 \le \\
       & \|{\bf DFE}(\lambda_{n_k})-{\bf DFE}(\bar \lambda)\|_2  + \|{\bf DFE}_{n_k}(\lambda_{n_k})- {\bf DFE}(\lambda_{n_k})\|_2
        \end{split}
    \label{inequality}
    \end{equation}
    Uniform convergence of ${\bf DFE}_{n_k}$ to ${\bf DFE}$ implies ${\bf DFE}$ is continuous. Hence, from (\ref{inequality}) we get ${\bf DFE}_{n_k}(\lambda_{n_k}) \to {\bf DFE}(\bar \lambda)$ and (\ref{weff}) implies
    \begin{equation}
    {\bf DFE}(\lambda)- {\bf DFE}(\bar\lambda) \not \in - {\rm int}\, \R^N, \ \forall \lambda \in \Lambda
        \label{weff1}
    \end{equation}
    i.e. $\bar \lambda \in {\rm WEff}({\bf DFE})$. 
    \item [ii)] Let $\lambda_{n}\in {\rm PEff}_C({\bf DFE}_{n})$ and let $\lambda_{n_k}$ be a subsequence converging to $\bar \lambda \in \Lambda$.  Then 
    \begin{equation}
    {\bf DFE}_{n_k}(\lambda)-{\bf DFE}_{n_k}(\lambda_{n_k}) \not \in -C\backslash \{0\}, \ \forall \lambda \in \Lambda
        \label{PEff-n}
    \end{equation}
    Passing to the limit we obtain 
    \begin{equation}
       {\bf DFE}(\lambda)-{\bf DFE}(\bar \lambda) \not \in -{\rm int}\, C, \ \forall \lambda \in \Lambda
    \end{equation}
    and hence 
    \begin{equation}
        {\bf DFE}(\lambda)-{\bf DFE}(\bar \lambda) \not \in-\R^n_+ \backslash\{0\}, \  \forall \lambda \in \Lambda
    \end{equation}
    \end{itemize}
\end{proof}

\begin{proof}
This is the proof of Proposition \ref{estimation}.
Let $\lambda(z^0) \in S^l_{z^0}$ be an isolated minimizer of order $\alpha$ and constant $m$ for $l(\cdot, z^0)$. Then, for  $\lambda (z) \in S_z^l$ it holds 
\begin{equation}
l(\lambda (z), z^0)-l(\lambda(z^0), z^0) \ge  h \|\lambda (z)- \lambda (z^0)\|^{\alpha}
\end{equation} 
We have 
\begin{equation}
l(\lambda(z^0), z)-l(\lambda(z), z)= l(\lambda (z^0), z^0)-l(\lambda (z) , z^0)+ w
\end{equation}
where 
\begin{equation}
w=[l(\lambda (z^0), z)-l(\lambda (z^0), z^0)]+ [l(\lambda (z), z^0)- l(\lambda (z), z)]
\end{equation}
We have 
\begin{equation}
|w|\le |l(\lambda (z^0), z)-l(\lambda (z^0), z^0)|+|l(\lambda (z), z^0)- l(\lambda (z), z)|\le 
\end{equation}
\begin{equation}
\sum_{i=1}^N\beta_i|g_i(\lambda (z^0), z_i)-g_i(\lambda (z^0), z_i^0)|+ \sum_{i=1}^N \beta_i|g_i(\lambda (z), z_i^0)- g_i(\lambda (z), z_i)|\le 
\end{equation}
\begin{equation}
2m \sum_{i=1}^N\beta_i d^Z(z_i, z_i^0)^{\delta}\le 2m \sum_{i=1}^N d^Z(z_i, z_i^0)^{\delta}
\end{equation}
We claim that 
\begin{equation}
l(\lambda (z), z^0) -l(\lambda (z^0) , z^0)\le |w|
\end{equation}
Indeed, suppose to the contrary that 
\begin{equation}
l(\lambda(z), z^0)-l(\lambda (z^0), z^0)) - |w|>0
\end{equation}
If $w=0$, then 
\begin{equation}
l(\lambda (z^0), z)- l(\lambda (z), z) < 0
\end{equation}
which contradicts $ \lambda (z) \in S_z^l$. If $w\not =0$, then we have
\begin{equation}
l(\lambda (z), z)-l(\lambda (z^0), z)=g(\lambda(z), z^0)-l(\lambda(z^0), z^0) - w>0
\end{equation}
which again contradicts  $ \lambda (z) \in S_z^l$. \ \\
Observe now that we have
\begin{equation}
h \|(\lambda (z)- \lambda (z^0)\| ^{\alpha}\le l(\lambda (z), z^0)-l(\lambda (z^0), z^0)
\end{equation}
and therefore 
\begin{equation}
h \|\lambda (z)- \lambda (z^0)\| ^{\alpha}\le l(\lambda (z), z^0)-l(\lambda (z^0), z^0)\le 2m \sum_{i=1}^N d^Z(z_i, z_i^0)^{\delta}
\end{equation}
So, it holds
\begin{equation}
\|\lambda (z)- \lambda (z^0) \|\le \left( \frac{2m}{h}\right)^{1/\alpha} \left(\sum_{i=1}^N d^Z(z_i, z_i^0)^{\delta}\right)^{1/\alpha}
\end{equation}
which finally implies
\begin{equation}
d(\lambda (z), S_{z^0}^l) \le \|\lambda (z)-\lambda (z^0)\| \le \left( \frac{2m}{h}\right)^{1/\alpha}\left(\sum_{i=1}^N d^Z(z_i, z_i^0)^{\delta}\right)^{1/\alpha}
\end{equation}
Since this holds for any $\lambda (z) \in S_z^l$  finally we have  
\begin{equation}
e(S_z^l, S_{z^0}^l) \le d(\lambda (z), S_{z^0}^l)\le \left( \frac{2m}{h}\right)^{1/\alpha}\left(\sum_{i=1}^N d^Z(z_i, z_i^0)^{\delta}\right)^{1/\alpha}
\end{equation}
Since   $\lambda(z) \in {\rm Eff}_z(DFE)$ we have  
\begin{equation}
d(\lambda(z), {\rm Eff}_{z^0}({\bf DFE}))\le \left( \frac{2m}{h}\right)^{1/\alpha}\left(\sum_{i=1}^N d^Z(z_i, z_i^0)^{\delta}\right)^{1/\alpha}
\end{equation}
which concludes the proof. 
\end{proof}

\begin{proof}
The proof of Proposition \ref{sta} is similar to that of Proposition \ref{estimation} and, therefore, it is omitted. 
\end{proof} 

\end{document}